\documentclass[sigconf]{acmart} 

\usepackage{algorithm}
\usepackage{algorithmic}
\AtBeginDocument{%
  }

\copyrightyear{2026}
\acmYear{2026}
\setcopyright{cc}
\setcctype{by}
\acmConference[RichMediaGAI '26]{The fourth International Workshop on Rich Media with Generative AI }{November 10--14, 2026}{Rio de Janeiro, Brazil}
\acmBooktitle{The fourth International Workshop on Rich Media with Generative AI (RichMediaGAI '26), November 10--14, 2026, Rio de Janeiro, Brazil}
\acmDOI{10.1145/3841458.3841545}
\acmISBN{979-8-4007-2947-8/2026/11}

\begin{document}

\title{Mechanistic Interpretability of Structure-Aware Numerical Reasoning in LLaMA 3.1–8B}

\author{Rahul Chowdhury}
\orcid{0009-0008-1575-6077}
\correspondingauthor
\affiliation{%
  \institution{Northeastern University}
  \city{Boston}
  \state{MA}
  \country{USA}
}
\email{chowdhury.rah@northeastern.edu}

\author{Timothy A Rupprecht}
\affiliation{%
  \institution{EmbodyX Inc.}
  \city{San Mateo}
  \state{CA}
  \country{USA}}
\email{tarupprecht@gmail.com}

\author{Senhao Cao}
\affiliation{%
  \institution{Northeastern University}
  \city{Boston}
  \state{MA}
  \country{USA}
}
\email{senhao.cao@gmail.com}
\author{Jiahao Liu}
\affiliation{%
 \institution{Mobi.ai}
 \city{Boston}
 \state{MA}
 \country{USA}}
\email{jiahao@takemobi.com}
\author{Octavia Camps}
\affiliation{%
  \institution{Northeastern University}
  \city{Boston}
  \state{MA}
  \country{USA}}
\email{O.Camps@northeastern.edu}

\author{David Bau}
\affiliation{%
  \institution{Northeastern University}
  \city{Boston}
  \state{MA}
  \country{USA}}
\email{davidbau@northeastern.edu}

\author{Pu Zhao}
\affiliation{%
  \institution{Northeastern University}
  \city{Boston}
  \state{MA}
  \country{USA}}
\email{p.zhao@northeastern.edu}
\author{Yanzhi Wang}
\affiliation{%
  \institution{Northeastern University}
  \city{Boston}
  \state{MA}
  \country{USA}}
\email{yanzhiwang@northeastern.edu}
\renewcommand{\shortauthors}{Rahul et al.}

\begin{abstract}
Recent work has shown that large language models (LLMs) 
exhibit strong numerical sequence modeling capabilities and show promise in time-series prediction. While 
LLMs display in-context learning capabilities, the mechanisms with which they accomplish 
time-series prediction remain unclear. Specifically, whether they truly understand the underlying structure, which at a minimum requires
reasoning over first differences in the sequence of numbers. 
To study this, we investigate Llama 3.1-8B 
from a mechanistic interpretability point of view. Mechanistic interpretability is an emerging field concerned with the reverse engineering of the algorithms learned by neural networks such as LLMs. To assess Llamas' numerical sequence modeling capabilities and to facilitate our mechanistic interpretability analysis, we create a sequence modeling task that cannot be solved without picking up structural cues. Specifically, we sample $n$ random numbers and repeat them with an offset. We find that Llama displays strong performance on our tasks suggesting that it can pick up on the underlying structure. To understand the mechanisms that allow it to do so, we perform probing experiments and activation patching based counterfactual analysis. Probing reveals that the model computes and stores first differences in its internal representations without explicit supervision, indicating that it tracks structural information about the sequence. Activation patching reveals that Llama retrieves the relevant first-difference with a mechanism similar to an induction circuit and subsequently adds it to the current value. Notably, our work represents one of the first studies to identify this form of concept induction in  LLMs. 
\end{abstract}




\maketitle

\section{Introduction}
\begin{figure}[t]
\centering
\includegraphics[width=0.95\columnwidth]{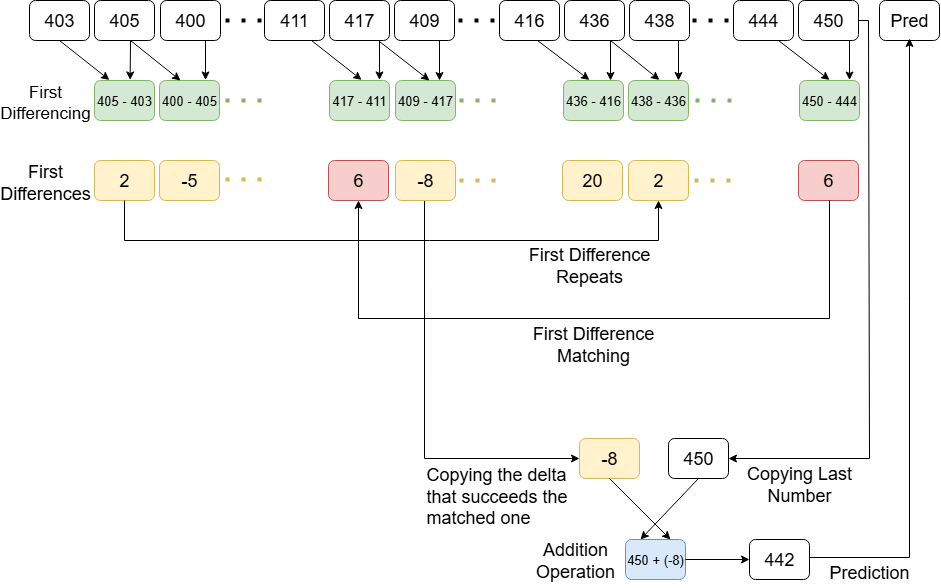} 
\caption{
Illustrates how the model identifies a repeated first-difference (delta) pattern after differencing, then extrapolates by locating and copying the delta that immediately follows the final observed one, and adding it to  last number. 
}
\label{fig:outline}
\end{figure}

Large Language Models (LLMs) can zero-shot extrapolate sequences, and  generate novel sequences and functions in response to complex queries \cite{mirchandani2023large,shen2024numerical,shen2024search,zhao-etal-2024-pruning,zhan-etal-2024-rethinking-token, bubeck2023sparks}. How do they do this? Did they encounter these trends during training and simply extrapolate using a rule learned specifically for those patterns? Or did they learn to infer the underlying structure of sequence data more generally?

If LLMs can understand the ordinal structure of numerical data — identifying a hidden pattern, reasoning about it, and applying a consistent extrapolation algorithm — the implications extend well beyond memorization. It would suggest these models are learning abstract structural reasoning, a capability with relevance far beyond language: reasoning over time-series data, understanding geometric relationships between image pixels, or maintaining structural coherence across connected points in 3D data \cite{shen2025draftattention,shen2025fastcar,shen2025efficient,shen2025quartdepth,shen2025sparse,shen2024lazydit,shen2024numerical,zhan2024exploring,zhan2024fast}.

\begin{figure*}[t]
\centering
\includegraphics[width=0.9\textwidth]{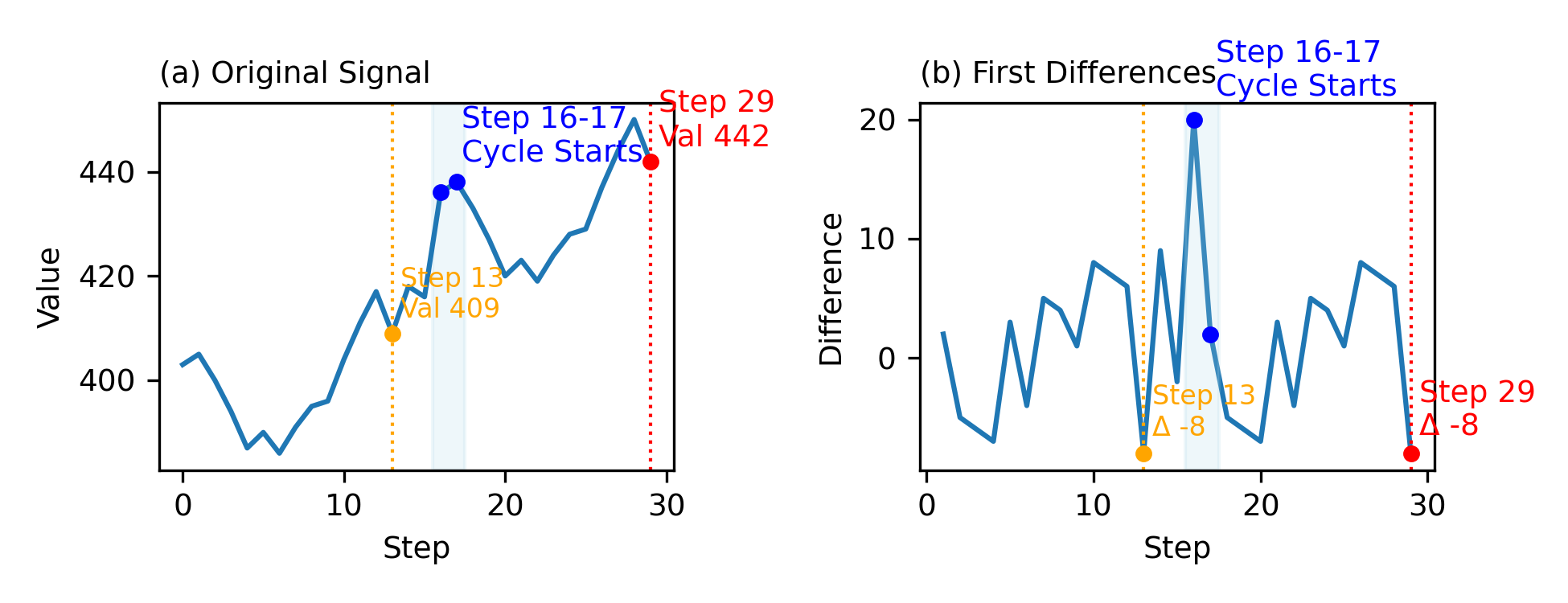} 
    \caption{
(a) Shows the input signal, which consists of all unique values which means no number appears twice in the sequence. (b) Shows the deltas of the input signal, revealing that a cycle or pattern emerges only after applying the differencing operation.  
}

\label{fig:wave-diff-comparison}
\end{figure*}

In this work, we study the LLaMA 3.1–8B model through the lens of mechanistic interpretability. To probe its structural reasoning capabilities, we introduce a non-trivial sequence modeling task, shown in Figure~\ref{fig:wave-diff-comparison}, that cannot be solved via token-level copying through induction heads \cite{elhage2021mathematical}. Our interpretability analysis combines linear probing with activation patching. Probing shows that first-difference representations are locally stored and linearly decodable across layers. Patching experiments further reveal that the model identifies structurally critical tokens, retrieves the correct offset through induction over these latent differences, and performs the final arithmetic step by adding that offset to the last number in the sequence. These computations are precise, localized, and structurally grounded.



To our knowledge, this is the first mechanistic interpretability study to uncover an internal mechanism by which an LLM performs induction over latent structure in numerical sequences—where extrapolation arises from arithmetic operations over internal representations retrieved through,  and we illustrate it in Figure~\ref{fig:outline}.
\section{Related Work}
\citet{mirchandani2023large} showed that LLMs can perform sequence completion tasks, attributing this ability to in-context learning. However, they did not examine whether such performance stems from memorization, nor did they investigate the underlying mechanisms or the model’s understanding of temporal structure. \cite{gruver2024large} proposed LLM-Time that explored ways to make a pre-trained LLMs fit for time-series forecasting, demonstrating that pretrained LLMs can perform well on time-series forecasting tasks without fine-tuning.  
Time-GPT \cite{garza2023timegpt} is a foundation model trained exclusively on time-series data, and Lag-Llama \cite{rasul2023lag} is another time-series-specific transformer. While these models highlight   effectiveness of transformers for sequence modeling \cite{zhao2024fully,zhan2021achieving,wu2022compiler,yang2023pruning,li2022pruning,rtseg,lin2025vote,10.1145/3746262.3761975,liu2025structured,guan2021cocopie,wang2018defending}, they fall outside  our study scope, as we focus on general-purpose LLM.

\citet{akyurek2022learning} and \citet{kantamneni2024transformers} examined in-context learning using toy models but did not explore whether LLMs can infer sequential structure in tasks they were not explicitly trained for. 
\cite{lan2023towards} applied mechanistic interpretability to identify shared circuits in simple sequence continuation tasks, such as extending short increasing number sequences. 
While  \cite{lan2023towards} demonstrated LLMs' ability to handle familiar patterns, it did not address whether these models can reason over more abstract or irregular numerical structures. In contrast, our work investigates whether a LLM can recognize and extrapolate more complex sequential structures—such as sequences defined by arbitrary first differences. This goes beyond surface-level extension and explores the model's ability to infer and extend abstract structure in number sequences.

\section{Problem Setting}

\subsection{Dataset}

We design a dataset to have strings containing numbers followed by a comma, and each instance finishes with a comma so that the predicted next token is a number. The numbers in the whole dataset were between 0 and 999. We construct each sequence with first differences from integers in the set containing all integers between $-9$ and $9$ inclusive, excluding $0$. Each sequence consists of unique numbers whose deltas are also unique up to the 17th position, after which the delta repeats.

We design the dataset to specifically test whether the LLM can uncover latent numerical structure and specifically track delta within a sequence.
As we illustrate in Figure~\ref{fig:wave-diff-comparison} and in an abbreviated form in Table~\ref{tab:wave-diff-comparison}, each sequence comprises two segments: the first is a random walk with unique deltas and no apparent pattern; the second segment reuses the same set of deltas, in the same order, as the first segment. This structure ensures that the model must recognize the pattern in delta and identify the unique location from which the model can copy the delta from.

\begin{table*}[t]
\centering
\renewcommand{\arraystretch}{1.3}
\begin{tabular}{c|*{22}{c|}c}
\hline
\textbf{Index} 
& 0 & 1 & 2 & 3 & 4 & 5
& 23 & 24 & 25 & 26 & 27
& 31 & 32 & 33 & 34 & 35 & 36 & 37
& 55 & 56 & 57 & \textbf{58} & 59 \\
\hline
\textbf{Token} 
& S & 403 & , & 405 & , & 400
& 411 & , & 417 & , & 409
& 416 & , & 436 & , & 438 & , & 433
& 444 & , & 450 & \textbf{,} & 442 \\
\hline
\textbf{Delta} 
&   &     &   & 2   &   & -5
& 7 &     & 6 &     & -8
& -2 &     & 20 &     & 2  &     & -5
& 7 &     & 6 &     & -8 \\
\hline
\end{tabular}

\caption{
Table shows selected index positions, tokens, and their corresponding first differences from the tokenized input. A recurrence in the first difference after token~33 marks the onset of a repeating cycle. 
}
\label{tab:wave-diff-comparison}
\end{table*}

\begin{algorithm}[tb]
\caption{Predict Next Number via First-Difference Pattern Detection}
\label{alg:predict-next}
{\raggedright
\textbf{Input}: Token sequence $S = [\text{BOS}, s_1, s_2, \ldots, s_n\texttt{,}]$\\
\textbf{Output}: Predicted next number $s_{n+1}$\par
}
\begin{algorithmic}[1]

\STATE Compute first differences:
\STATE \hspace{1em} \parbox[t]{.85\linewidth}{$D = [s_2 - s_1,\ s_3 - s_2,\ \ldots,\ s_n - s_{n-1}]$}
\STATE Identify repeating pattern $P$ in $D$
\STATE Locate the index $k$ where $P$ first appears in $D$
\STATE Determine the phase position $p$ of $s_n$ within pattern $P$
\STATE Let $d_p = D[p]$ be the corresponding first difference
\STATE Compute prediction: $s_{n+1} = s_n + d_p$
\STATE \textbf{return} $s_{n+1}$
\end{algorithmic}
\end{algorithm}
\subsection{Model}
We conduct all experiments using LLaMA 3.1–8B \cite{grattafiori2024llama3herdmodels}, a transformer \cite{vaswani2017attention} based model with 32 layers and 32 attention heads per layer. One key reason for choosing this model is that its tokenizer represents each integer from 0 to 999 as a single token. This property makes the analysis more tractable by reducing the number of tokens per sequence, thereby simplifying both intervention and observation during interpretability experiments. As a result, LLaMA 3.1–8B was used consistently across all experiments.

\subsection{Software}
 We extensively use NNsight and NDIF framework~\cite{fiottokaufman2024nnsightndifdemocratizingaccess} for tracing activations, applying interventions, and analyzing internal representations of the LLaMA~3.1–8B model. 

\subsection{Performance Evaluation}
We evaluate LLaMA 3.1-8B on the sequence prediction task as shown in Figure~\ref{fig:outline} and Figure~\ref{fig:wave-diff-comparison}, and we  formally describe it in Algorithm~\ref{alg:predict-next}. We prompt the model to predict the 30\textsuperscript{th} number given the first 29 elements of each input sequence. The evaluation dataset consists of 10{,}000 sequences, 
each formatted as a comma-separated string of integers. Every instance ends with a comma, indicating that the next token to be predicted should be a number.

Due to this formatting, we tokenize each number and each comma separately, as we  illustrate in abbreviated form in Table~\ref{fig:wave-diff-comparison}. Including the initial special token, the model processes 59 tokens before generating the 60\textsuperscript{th} token as output. This prediction corresponds to the 30\textsuperscript{th} number in the sequence.

\subsubsection{Metrics}
We employ two complementary evaluation metrics:

\textit{Mean Absolute Error (MAE)} quantifies average prediction error:
\begin{equation}
\text{MAE} = \frac{1}{n} \sum_{i=1}^{n} \left| y_i - \hat{y}_i \right|,
\end{equation}
where $y_i$ and $\hat{y}_i$ denote the ground truth and predicted values, respectively, and $n$ is the number of instances.

\textit{Coefficient of Determination ($R^2$)} measures explained variance:
\begin{equation}
R^2 = 1 - \frac{\sum_{i=1}^{n} (y_i - \hat{y}_i)^2}{\sum_{i=1}^{n} (y_i - \bar{y})^2},
\end{equation}
where $\bar{y}$ is the mean of ground truth. Values close to 1 indicate strong predictive performance, while values near 0 or negative suggest performance close to or worse than a naive mean predictor.

\subsubsection{Results}
LLaMA 3.1-8B achieves an MAE of 4.2748 and an $R^2$ of 0.9958 on our evaluation dataset. The high $R^2$ value indicates that the model explains approximately 99.58\% of the variance in the target sequences, demonstrating strong pattern recognition capabilities for numerical sequence prediction.




\section{Is Numerical Information Represented Locally in the Model?}

\label{table:three-row-final}

\begin{figure*}[t]
\centering
\includegraphics[width=0.8\textwidth]{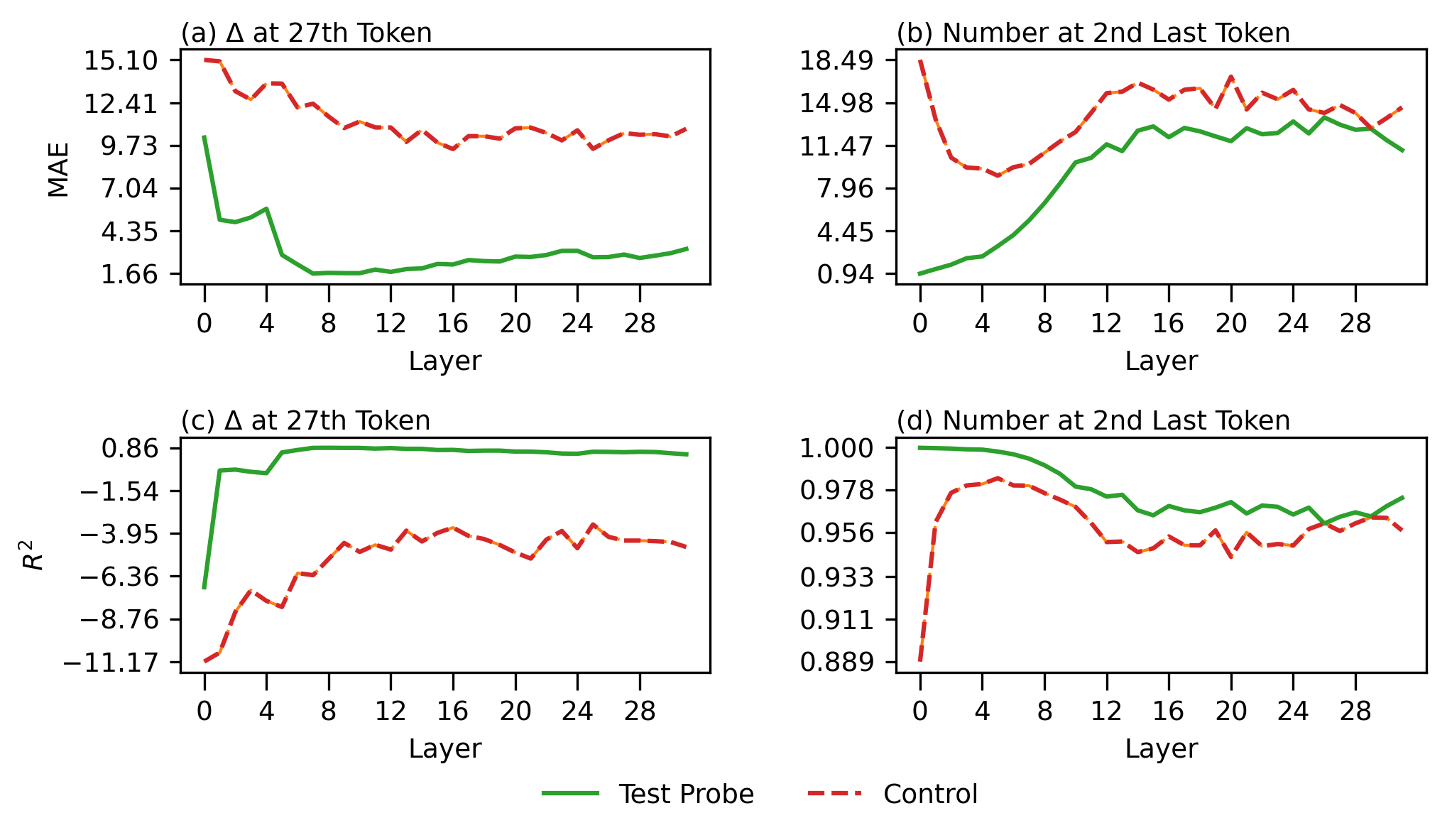} 

\caption{Layerwise linear decodability of first differences and final numbers. Mean absolute error (MAE; top) and coefficient of determination ($R^2$; bottom) for linear probes trained to decode delta (left) at 27\textsuperscript{th} position and final number (right) at 57\textsuperscript{th} position from hidden states across layers. Solid green lines denote performance on true targets; dashed red lines denote control probes trained on shuffled labels. Delta show consistently strong linear decodability, while final number representations gradually lose decodability. }

\label{fig:probing}
\end{figure*}

To predict the next number, the model must add the final observed value to a previously seen first difference. This raises a core interpretability question: does the model explicitly encode these numerical components in structured, localized representations, or does it rely solely on implicit pattern recognition?

We use \textbf{probing} experiments to assess the representational content of the model’s hidden states. Unlike patching—which tests whether an activation is \emph{functionally necessary} by measuring how altering it affects the model’s output—probing asks whether specific information is \emph{present} in the representation, regardless of whether the model ultimately uses it. Here, we probe for the presence and location of numerically meaningful quantities in the hidden layers by training linear regressors to decode two key values: (1) the first difference from the 27\textsuperscript{th} token position (where it is expected to be stored), and (2) the final number from the 57\textsuperscript{th} token position. If successful, this would indicate that the model explicitly encodes these arithmetic components in localized, interpretable forms, even without task-specific supervision, and uses these interpretable representations for prediction.
\subsection{Probing Setup}

We collect hidden states from two key token positions: the 27\textsuperscript{th} token position, which is expected to encode the first difference that must be added to the final number to predict the next number, and the 57\textsuperscript{th} token position, which contains the final number in the input sequence. At each layer, we train linear regression models using ordinary least squares on 4{,}500 training samples and 500 held-out samples for testing:
\begin{equation}
\small
\hat{\boldsymbol{\beta}} = \operatorname*{arg\,min}_{\boldsymbol{\beta}} \left\| \mathbf{Y} - \mathbf{X} \boldsymbol{\beta} \right\|^2,
\end{equation}
where $\mathbf{X} \in \mathrm{R}^{n \times d}$ is the matrix of hidden states (with $d = 4096$ dimensional representations across $n = 4500$ training examples), $\mathbf{Y} \in \mathrm{R}^n$ is the target vector (either the first difference or the final number), and $\boldsymbol{\beta} \in \mathrm{R}^d$ is the learned weight vector.

To establish baselines, we implement control conditions that test whether observed decodability reflects genuine localization or spurious correlations. For first difference decoding, we train control probes using activations from the 57\textsuperscript{th} token to predict the first difference of 27\textsuperscript{th} position, testing whether first difference information is accessible from positions nearer to the last token position. For final number decoding, we train control probes using activations from the 27\textsuperscript{th} token to predict the number at the 57\textsuperscript{th} position. These  ensure that successful decoding reflects position-specific encoding rather than global numerical information distribution.

\subsection{Results}

Figure~\ref{fig:probing}  presents the results of  probing experiments. The top row reports the MAE of the linear regression model on held-out test samples, while the bottom row shows the $R^2$. We compute these metrics at each layer to evaluate the linearity of the target quantities throughout the model's depth.

The results confirm that both the first difference (at the 27\textsuperscript{th} token position) and the final number (at the 57\textsuperscript{th} token position) are linearly decodable from the hidden states, as evidenced by low MAE values and high $R^2$ scores approaching 1. Notably, the first difference becomes linearly accessible starting after the 7\textsuperscript{th} layer, where the MAE drops sharply and approaches its minimum, while the $R^2$ score increases and stabilizes above 0.8. Although performance slightly decays with depth, it remains relatively stable across the first 16 layers.
In contrast, the linear decodability of the final number at the 57\textsuperscript{th} position shows a gradual decline across deeper layers, with MAE increasing and $R^2$ decreasing after layer 8. This suggests that the final number representation 
becomes increasingly entangled with other contextual information as it propagates through  model.

Crucially, the control conditions demonstrate substantially worse performance across all layers for both targets. For first differences, control probes using 57\textsuperscript{th} token activations achieve negative $R^2$ values and high MAE, indicating no meaningful linear relationship. Similarly, control probes attempting to decode final numbers from 27\textsuperscript{th} token activations show consistently poor performance. This confirms that the observed decodability is position-specific and not due to general numerical information 
in the representations.

Together, these results indicate that the numerical components necessary for the model's arithmetic composition-the first difference and the final number-are linearly encoded in specific  locations. 

\section{Is the Next Number Computed by Adding a Retrieved First Difference?}
\begin{figure*}[t]
\centering
\includegraphics[width=0.8\textwidth]{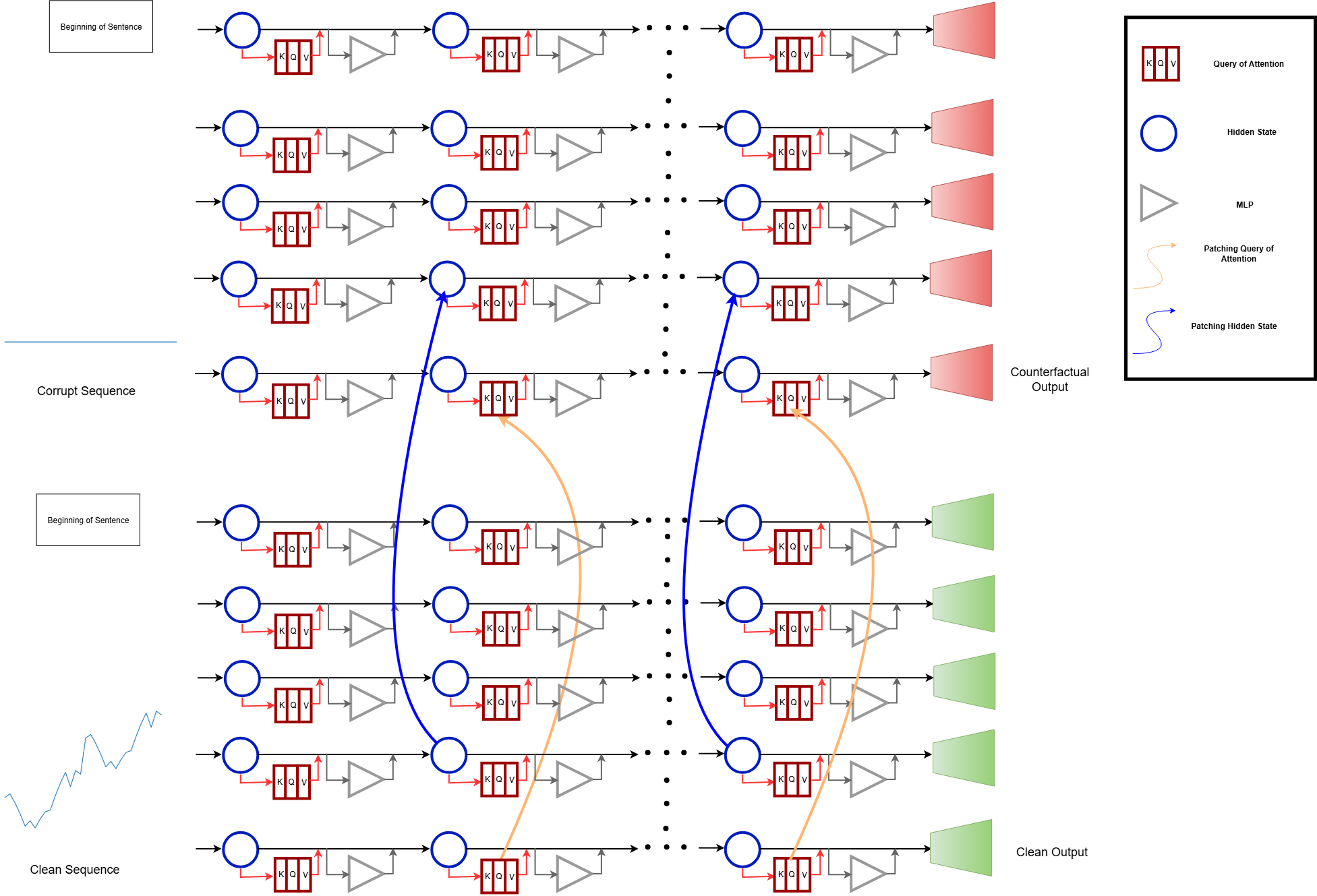} 
\caption{
Illustrates the query-hidden state patching technique that we use to test whether the model performs arithmetic composition by retrieving a first difference from an earlier position. The clean sequence (bottom) contains a structured delta pattern, while the corrupt sequence (top) is flat-valued. We transplant activations from the clean run into the corrupt run from a given layer $l$ onward. We patch hidden states (shown with blue arrows) at layers $l$ to $L$ at one token position at a time except the beginning-of-sequence and final two tokens. We patch the query vector of the final token (shown with orange arrows) from the clean run at layers $l+1$ to $L$, ensuring the model attends to the correct location encoding the target first difference. This setup isolates and tests the causal contribution of induction and addition behaviors in the model's prediction pipeline.
}
\label{fig:Q-Hidden}
\end{figure*}

Having established that the model encodes first differences and final numbers in localized representations, we now ask whether it \emph{functionally uses} those representations to perform arithmetic composition. We test whether the model retrieves a previously seen first difference from a particular token position and perform arithmetic operation on it.

To answer this question, we use \textbf{activation patching}
\cite{wang2023interpretability, meng2022locating,vig2020investigating,zhao2025taming,zhao2026hieramp,zhao2026omnimem,goldowsky2023localizing,finlayson2021causal}, following the patching methodology of \cite{dumas2024llamas}. We transplant activations from a clean sequence into a corrupt one to determine if the model retrieves the delta from the hypothesized position (delta that follows the latest one) and adds that to a new number of an unseen sequence.

\subsection*{Clean vs. Corrupt Sequence}

We define two types of input sequences:
\begin{itemize}
    \item \textbf{Clean sequence}: $x^{\text{clean}} = (x_1, \dots, x_{29})$, where the deltas $\Delta_i$ begin to repeat after $i = 17$, and the model predicts:$x_{30} = x_{29} + \Delta^*$ where $\Delta^*$ is the difference retrieved from an earlier position. In tokenized form, $x_{29}$ appears at token 57, and the output $x_{30}$ is generated at token 58.
    \item \textbf{Corrupt sequence}: $x^{\text{corrupt}} = (100, 100, \dots, 100)$, so that $\Delta_i = 0$ for all $i$.

\end{itemize}

What makes the \textit{corrupt sequence} particularly well-suited for evaluating the extraction and addition of the first difference is its deliberately flattened structure, which eliminates any intrinsic gradient or variation in the input. The corrupt sequence serves as an ideal control, as it lacks the algorithm present in the clean sequence, allowing us to test whether the model can still retrieve and apply a meaningful first difference from an alternate source.

\subsection{Patching Setup}

Let $H_t^{(l)}$ and $Q_t^{(l)}$ denote the hidden state and query vector at token $t$, layer $l$. We patch hidden states to all tokens 
except  the beginning 
token, the final number token (57), and the final token (58):
\[
H_t^{\text{corrupt},(l:L)} \leftarrow H_t^{\text{clean},(l:L)}
\]
The query vector of  final token (58) is patched from one layer above:
\[
Q_{58}^{\text{corrupt},(l+1:L)} \leftarrow Q_{58}^{\text{clean},(l+1:L)}
\]

In particular, following \cite{dumas2024llamas}, we patch layer $l$ and all subsequent layers,  as we illustrate in Figure~\ref{fig:Q-Hidden}. Patching downstream layers ensures that transient causal signals are preserved and not suppressed as noise by later layers, thereby, improving sensitivity to the influence of critical nodes on the final prediction. 

For the final token, we specifically patch the query vector across layers $l+1$ through $L$. This is necessary because the corrupt sequence does not learn in-context to retrieve target first differences from earlier tokens. By patching the query vector of the final token from the clean forward pass, we ensure that the model attends to the position encoding the appropriate first difference, thereby, facilitating the intended retrieval behavior.
\subsection{Causal Metric}

To quantify causal influence, we use   difference in predicted probabilities of   \textit{counterfactual label}  $y_{\text{cf}} = 100 + \Delta^*$ as   evaluation metric:
\begin{equation}
\Delta P = \mathrm{E}_{x \sim \mathcal{D}} \left[ P_{\text{patched}}(y_{\text{cf}} \mid x) - P_{\text{corrupt}}(y_{\text{cf}} \mid x) \right]
\end{equation}
where $P_{\text{patched}}(y_{\text{cf}} \mid x)$ is the probability assigned to the \textit{counterfactual label} $y_{\text{cf}}$ under the patched forward pass, and $P_{\text{corrupt}}(y_{\text{cf}} \mid x)$ is the corresponding probability under the corrupt forward pass. A higher $\Delta P$ indicates stronger causal contribution of the patched components to the model's ability to predict the correct label.

\subsection{Results}

\begin{figure}[t]
\centering
\includegraphics[width=0.9\columnwidth]{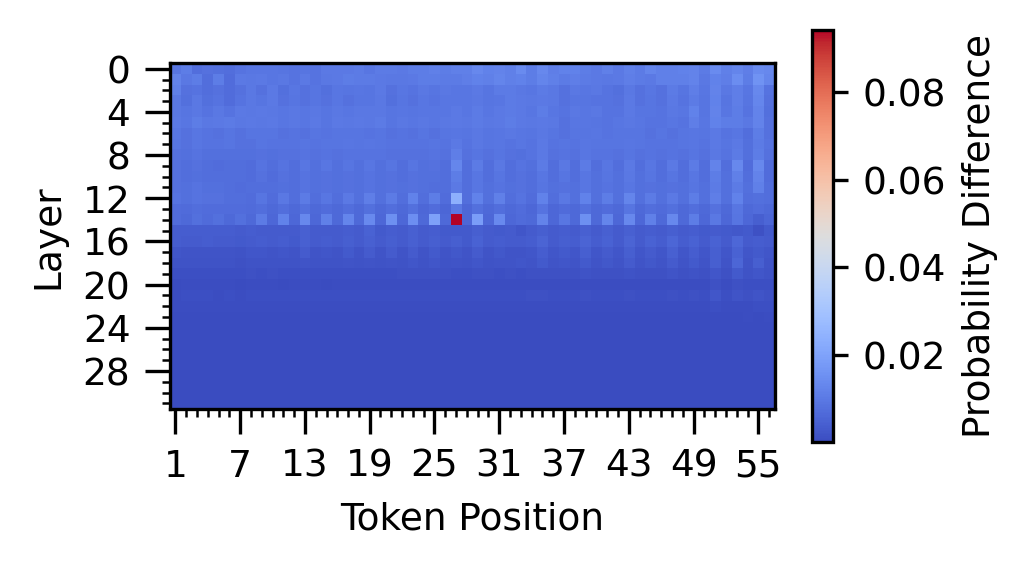} 
\caption{
Heatmap of probability difference scores ($\Delta P$) across layers and token positions, following patching of the final query vector from layers $l{+}1$ to $L$ and hidden states from layers $l$ to $L$. A sharp causal effect emerges at token~27, highlighted in red, indicating that the final query attends to position~27 to retrieve the locally stored first difference that follows the last observed one.
}

\label{fig:arithmeticcomp}
\end{figure}

Figure~\ref{fig:arithmeticcomp} presents the results of the patching experiment conducted on 100 data instances with zero MAE to ensure a strong response signal. We observe a significant increase in the probability difference score ($\Delta P$), highlighted in red in Figure~\ref{fig:arithmeticcomp}, at the 27\textsuperscript{th} token position, beginning from layer~14 onward. This indicates that, from layer~15 onward, the query vector of the final token (position~58) causally interacts with the hidden state at position~27. The arithmetic operation is highly localized: only at these layers does the model cleanly separate the corrupt final number and compose it with the correct first difference. Earlier interventions disrupt this alignment due to representational entanglement between the corrupt and patched sequences, which impairs computation.

This finding supports the hypothesis that the model performs induction over first differences by locating and copying a locally stored delta representation—analogous to the [A][B]...[A][?] pattern discussed in \cite{elhage2021mathematical}—that follows the final observed one, though this behavior operates at a structural rather than token level. The arithmetic operation that follows induction—adding the copied delta to the final number at position~57 to generate the next token—is generalizable, as the model performs it even in the absence of a guiding algorithm in the corrupt sequence. This confirms that the model engages in both induction and arithmetic composition in numerical sequence extrapolation.

\section{Can the Model Identify Functionally Critical Tokens for Delta Retrieval and Extrapolation?}

Although previous experiments demonstrate that the model stores first differences in localized representations and retrieves the appropriate ones for arithmetic composition, they do not tell if the model   uncovers the underlying algorithm. In particular, uncovering requires identifying the onset of a repeating pattern, aligning it with the final observed difference, and selecting the correct delta to apply. 
To investigate this, we use patching to isolate the causal contribution of each attention head and assess which heads drive successful extrapolation by routing value from   appropriate positions.

\subsection*{Patching Setup}
We use the same clean and corrupt sequences described in the previous section, where the clean input contains a repeating first-difference pattern and the corrupt input is a flat sequence of 100s. The target counterfactual label remains: $y_{\text{cf}} = 100 + \Delta^*$.

Let $Q_s^{(l,h)}$, $K_t^{(l,h)}$, $V_t^{(l,h)}$, and $W_O^{(l,h)}$ denote the query, key, value vectors, and output projection matrix for head $(l,h)$ in layer $l$ at token $s$ attending to token positions $t$. The attention weights at head $(l,h)$ are computed as:
\[
\alpha_{t,s}^{(l,h)} = \mathrm{softmax}\left( \frac{(Q_s^{(l,h)})^\top K_t^{(l,h)}}{\sqrt{d_h}} \right)
\]

The corresponding output of attention head $(l,h)$ at token $t$ is:
\[
z_s^{(l,h)} = W_O^{(l,h)} \left( \sum_{t=1}^{T} \alpha_{t,s}^{(l,h)} \cdot V_t^{(l,h)} \right)
\]

To evaluate the causal effect of head $(l,h)$ at the final token position $s = 58$, we patch just its output, as we illustrate in Figure~\ref{fig:patchingheads} in the appendix , from the clean run into the corrupt run:
\[
z_{58}^{\mathrm{corrupt},(l,h)} \leftarrow z_{58}^{\mathrm{clean},(l,h)}
\]


This setup isolates the contribution of a single attention head by substituting its output at the final token while preserving the rest of the corrupt context.

\subsection{Causal Metric}

To determine which tokens most significantly contribute to the model’s prediction, we construct a profile of attention heads that integrates two key signals: (1) the \textit{causal effect} of each head, and (2) the \textit{value-weighted attention mass} it allocates across token positions. We compute the causal effect as the change in the model’s predicted probability for the counterfactual label when we patch head’s output at position $s = 58$ from the clean to the corrupt run, as we describe in the previous section. The second component—the value-weighted attention— identifies how much of its value is added to the residual stream through this head.

We define the token importance score $\mathrm{TI}_t$ as:
\begin{equation}
\small
\mathrm{TI}_t = \frac{1}{N} \sum_{n=1}^{N} \lambda_n^{(l,h)} \cdot w_{n,t}^{(l,h)},
\label{eq:token-importance}
\end{equation}
where  $N$ is the number of evaluation samples, $\lambda_n^{(l,h)} = P_{\mathrm{patched}}(y_{\mathrm{cf}} \mid x_n) - P_{\mathrm{corrupt}}(y_{\mathrm{cf}} \mid x_n)$ is the causal effect of patching head $(l,h)$ in sample $n$, and  $w_{n,t}^{(l,h)}$ is the value-weighted attention score at the attended position $t$.
The value-weighted attention score $w_{n,t}^{(l,h)}$ is defined using the attention weights and value norms:
\begin{equation}
w_{n,t}^{(l,h)} = \alpha_{t,s}^{(l,h)} \cdot \left\| V_t^{(l,h)} \right\|_2,
\label{eq:value-weighted}
\end{equation}

This metric $\mathrm{TI}_t$ weights causal influence with value-aware attention, providing a principled measure of token-level functional relevance to the model's output.

\subsection{Results}
\begin{figure}[t]
\centering
\includegraphics[width=0.95\columnwidth]{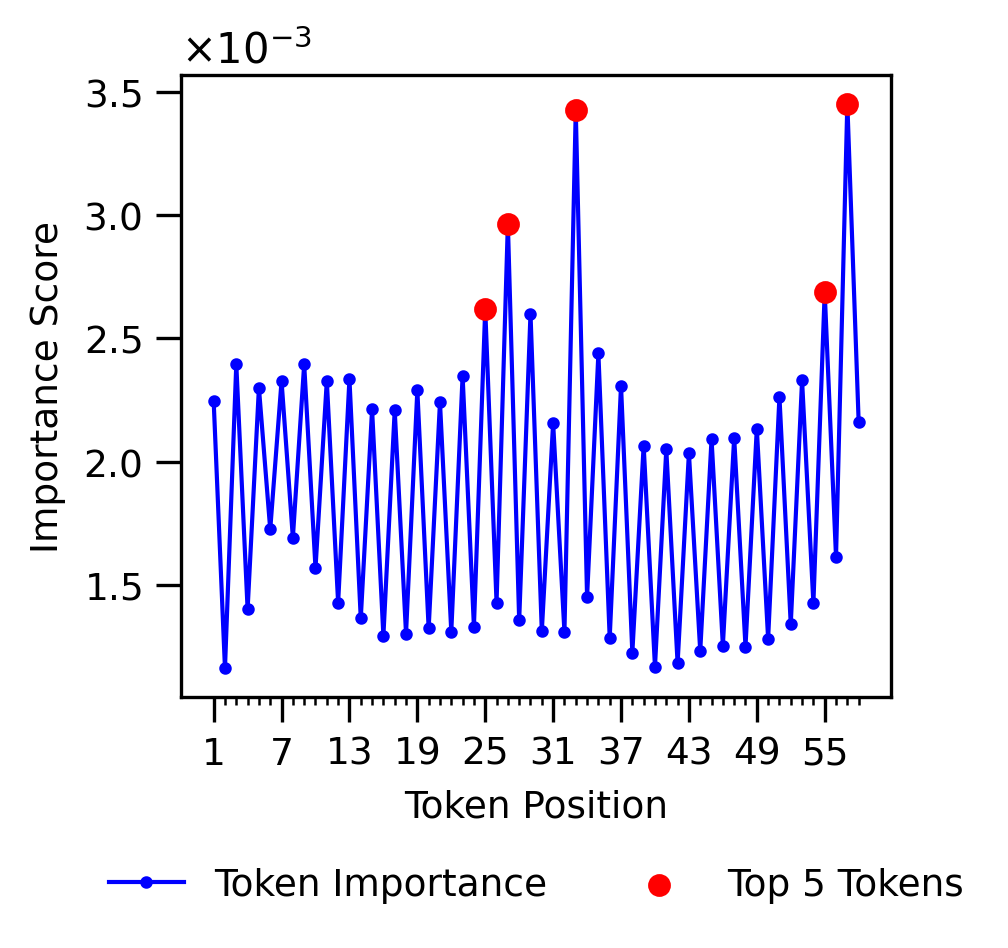} 
\caption{
Token importance scores ($\mathrm{TI}_t$) across sequence positions. Peaks highlight the model's use of structural alignment for pattern detection and delta retrieval.
}

\label{fig:ti}
\end{figure}

Figure~\ref{fig:ti} presents the results of the causal attention-based analysis over 100 sequences with zero MAE, and the top five most relevant tokens  are identified as positions 57, 33, 27, 55, and 25. As we show in Table~\ref{tab:wave-diff-comparison}, these correspond to the critical points that the model needs for identification and extrapolation of the pattern: the final number (57), the onset of the repeating delta pattern (33), the delta that gets added to the final value (27), the number that the model needs to compute the last delta (55), and a token (25) that both matches the final delta and enables retrieval of the next one.
 
These results also suggest the model emphasizes tokens that support first-difference alignment, with token 27 playing a key role in extrapolation. The pattern of attention supports the hypothesis that the model recognizes the pattern and must have first used the final first difference as a retrieval anchor to locate the appropriate delta, and, upon recognizing the repeating structure, may further leverage phase alignment and positional encoding to guide delta selection.

\section{Does the Model Prioritize Delta Induction Over Positional Cues?}
Prior experiments demonstrate that the model stores and composes first differences to extrapolate future values. However, they do not reveal  whether retrieval is governed by  semantic identity of  delta or by its positional and phase-based alignment in  input.

To investigate this, we perform a key swapping intervention between the 27\textsuperscript{th} and 25\textsuperscript{th} tokens—two positions with high and comparable importance scores as shown in Figure~\ref{fig:ti}.
\subsection{Rotary Position Embedding (RoPE)}

Let $x \in \mathbf{R}^d$ denote an input vector, where $d$ is even. Following the formulation introduced by~(Su et al. 2024)\cite{su2024roformer}, we partition $x$ into $d/2$ adjacent 2D pairs:
\[
x = \left[\, (x_1^{(1)}, x_1^{(2)}),\, (x_2^{(1)}, x_2^{(2)}),\, \dots,\, (x_{d/2}^{(1)}, x_{d/2}^{(2)}) \,\right]
\]

Let $W_q, W_k \in \mathbf{R}^{d \times d}$ be the query and key projection matrices. For position $m$, the RoPE-transformed query is:
\[
Q_m = \text{RoPE}(W_q x, m) = \bigoplus_{i=1}^{d/2} R(m\theta_i)\, (W_q x)_i
\]
and similarly, for position $n$, the RoPE-transformed key is:
\[
K_n = \text{RoPE}(W_k x, n) = \bigoplus_{i=1}^{d/2} R(n\theta_i)\, (W_k x)_i
\]

Here, $R(m\theta_i)$ is a 2D rotation matrix applied to the $i$-th 2D component:
\[
R(m\theta_i) =
\begin{bmatrix}
\cos(m\theta_i) & -\sin(m\theta_i) \\
\sin(m\theta_i) & \cos(m\theta_i)
\end{bmatrix}
\]

Here, $R(n\theta_i)$ is a 2D rotation matrix applied to the $i$-th 2D component:
\[
R(n\theta_i) =
\begin{bmatrix}
\cos(n\theta_i) & -\sin(n\theta_i) \\
\sin(n\theta_i) & \cos(n\theta_i)
\end{bmatrix}
\]
The operator $\bigoplus$ denotes vector concatenation across the $d/2$ rotated 2D components:
\[
\bigoplus_{i=1}^{d/2} v_i = [v_1;\, v_2;\, \dots;\, v_{d/2}]
\]
where each $v_i \in \mathbf{R}^2$, yielding a final vector in $\mathbf{R}^d$.

The query and key vectors are independently rotated using their respective absolute positions $m$ and $n$, and the relative positional information gets encoded through their dot product $Q_m^\top K_n$.
\subsection{Patching Setup}
LLaMA 3.1-8B uses rotary position embedding (RoPE) \cite{su2024roformer} where positional information is local to key and query. 
We swap Keys $K_{27}^{\text{clean},(l:L)} \leftrightarrow K_{25}^{\text{clean},(l:L)}$ in layer $l$ and all subsequent layers to redirect positional and phase-based cues while preserving $V_{27}^{\text{clean},(l:L)}$ and $V_{25}^{\text{clean},(l:L)}$ by patching the values from the original run to maintain their first difference content across layers $l+1$ through $L$, as we illustrate in Figure~\ref{fig:swapkeys}  in the appendix. This setup attempts to disentangle two types of cues bound to the delta representation.

\subsection{Causal Metric}
To quantify the causal impact of an intervention, we measure the change in the model’s confidence for both the counterfactual label \( y_{\text{cf}} = x_{29} + \Delta_{25} \) and the ground truth label \( y_{\text{gt}} = x_{29} + \Delta_{27} \). The causal effect on the counterfactual label is defined as:
\begin{equation}
\Delta P_{\text{cf}} = \mathrm{E}_{x \sim \mathcal{D}} \left[ P_{\text{patched}}(y_{\text{cf}} \mid x) - P_{\text{corrupt}}(y_{\text{cf}} \mid x) \right],
\end{equation}
and for the ground truth label:
\begin{equation}
\Delta P_{\text{gt}} = \mathrm{E}_{x \sim \mathcal{D}} \left[ P_{\text{patched}}(y_{\text{gt}} \mid x) - P_{\text{corrupt}}(y_{\text{gt}} \mid x) \right].
\end{equation}

We also report the absolute probabilities assigned  after the intervention: $P_{\text{patched}}(y_{\text{cf}} \mid x)$ and $P_{\text{patched}}(y_{\text{gt}} \mid x)$.
These metrics together capture both the direction and magnitude of the model’s response to the intervention.

\subsection{Results}

\begin{figure}[t]
\centering
\includegraphics[width=0.99\columnwidth]{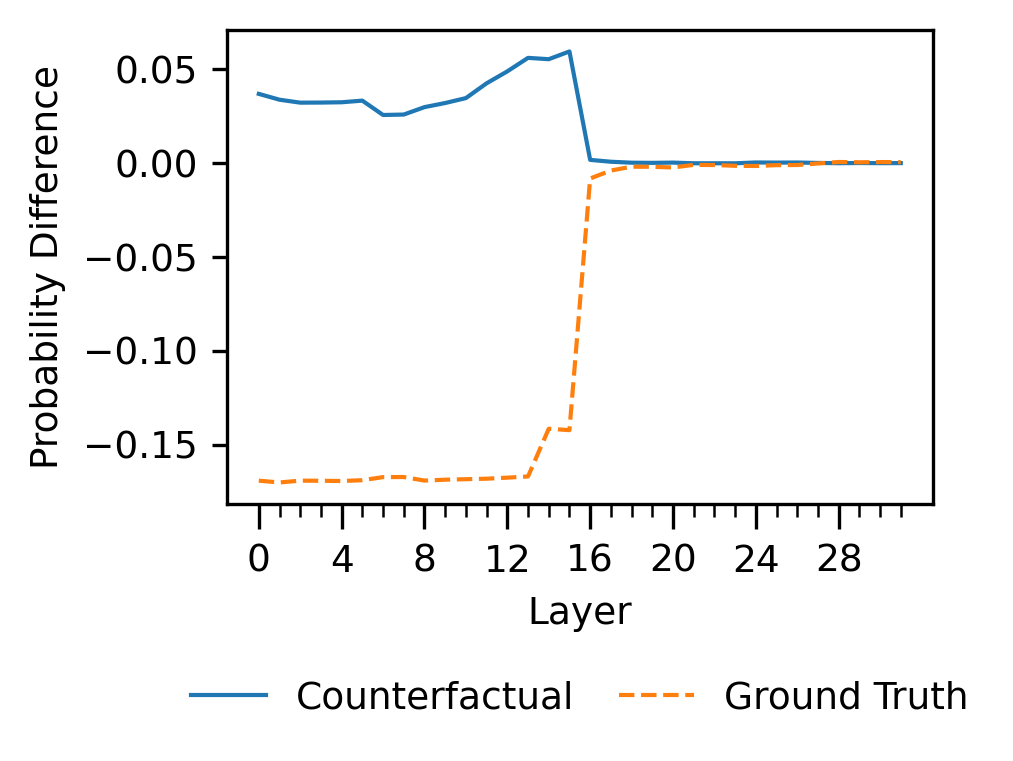} 
\caption{Shows key-swap intervention shifts confidence from ground truth to counterfactual, revealing the role of position-aware attention alignment in delta retrieval.}

\label{fig:probdiffkeyswap}
\end{figure}

\begin{figure}[t]
\centering
\includegraphics[width=0.99\columnwidth]{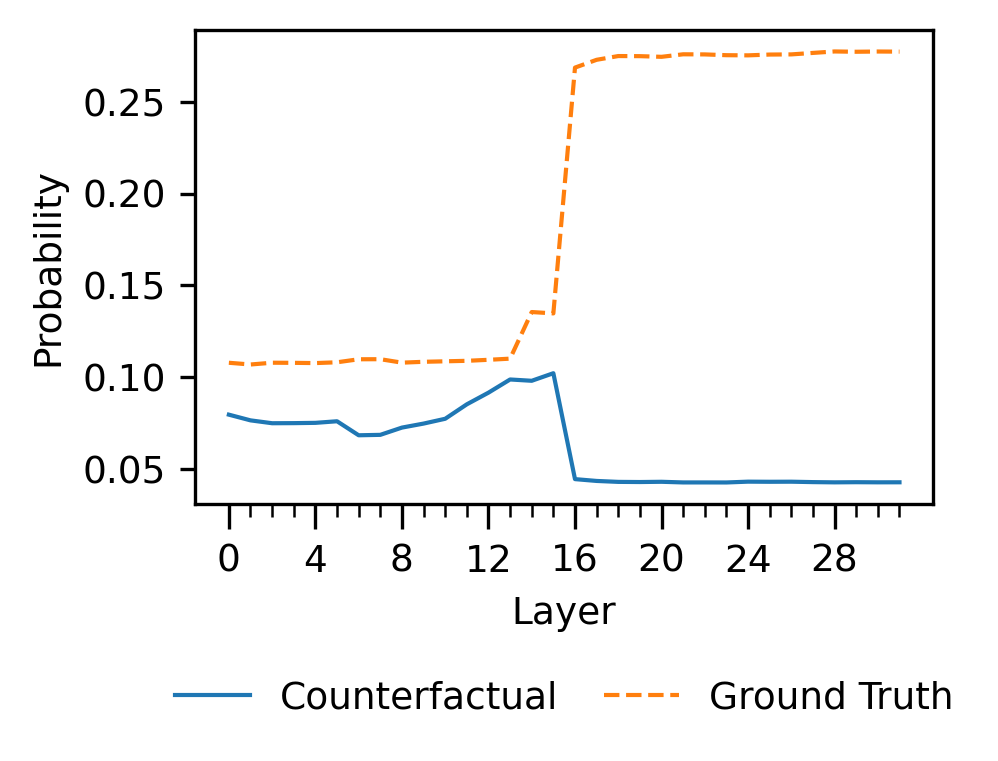} 
\caption{Shows ground truth remains more probable post key-swap, indicating the model prioritizes delta-based composition over redirected attention.}
\label{fig:probkeyswap}
\end{figure}

Figure~\ref{fig:probdiffkeyswap} presents the probability difference following the key-swapping intervention $K_{27}^{\text{clean},(l:L)} \leftrightarrow K_{25}^{\text{clean},(l:L)}$ that we conduct over 100 instances with zero MAE. The intervention leads to a significant increase in \( \Delta P_{\text{cf}} \) and a corresponding drop in \( \Delta P_{\text{gt}} \), suggesting that phase or position-based cues contribute meaningfully for delta retrieval and composition. However, as shown in Figure~\ref{fig:probkeyswap}, which reports the absolute probabilities after intervention, the drop in \( P_{\text{patched}}(y_{\text{gt}}) \) was not sufficient for \( P_{\text{patched}}(y_{\text{cf}}) \) to overtake it. This indicates that although key-based redirection influences the model’s behavior, it is ultimately not strong enough to override the original delta-based composition since the value vector must contain information that this first difference comes after the last observed one. The model’s affinity for delta-based composition supports the hypothesis that it first performed induction over first differences to learn the underlying structure: internally simulated the composition process, and identified the correct algorithm, potentially refining it through phase-based alignment.

\section{Conclusion}

This work demonstrates the LLM, despite not being explicitly trained for high-precision numerical tasks, exhibit an emergent ability to perform structural extrapolation in numerical sequences without any explicit supervision. We show that the LLM is capable of recognizing trends, inferring position-dependent rules encoded as distinct first differences, and systematically composing these rules to generate accurate predictions over evolving input sequences. Our analyses reveal that the LLM exhibit reasoning behaviors, operating over latent structure. Patching provided strong evidence that the model identifies numerical patterns, computes first differences even before recurring structures emerge, stores these differences locally, and later retrieves them to generate accurate predictions. Probing confirmed that the observed effects are not incidental but reflect functional computations over structured internal representations.

We believe these findings would invite future work on extending these insights to other reasoning tasks where latent structure plays a critical role.  

\bibliographystyle{ACM-Reference-Format}
\bibliography{sample-base}

\newpage
\onecolumn
\appendix

\section*{Appendix}
\section{Illustrating Patching Experiments}
\begin{figure*}[h]
\centering
\includegraphics[width=0.9\textwidth]{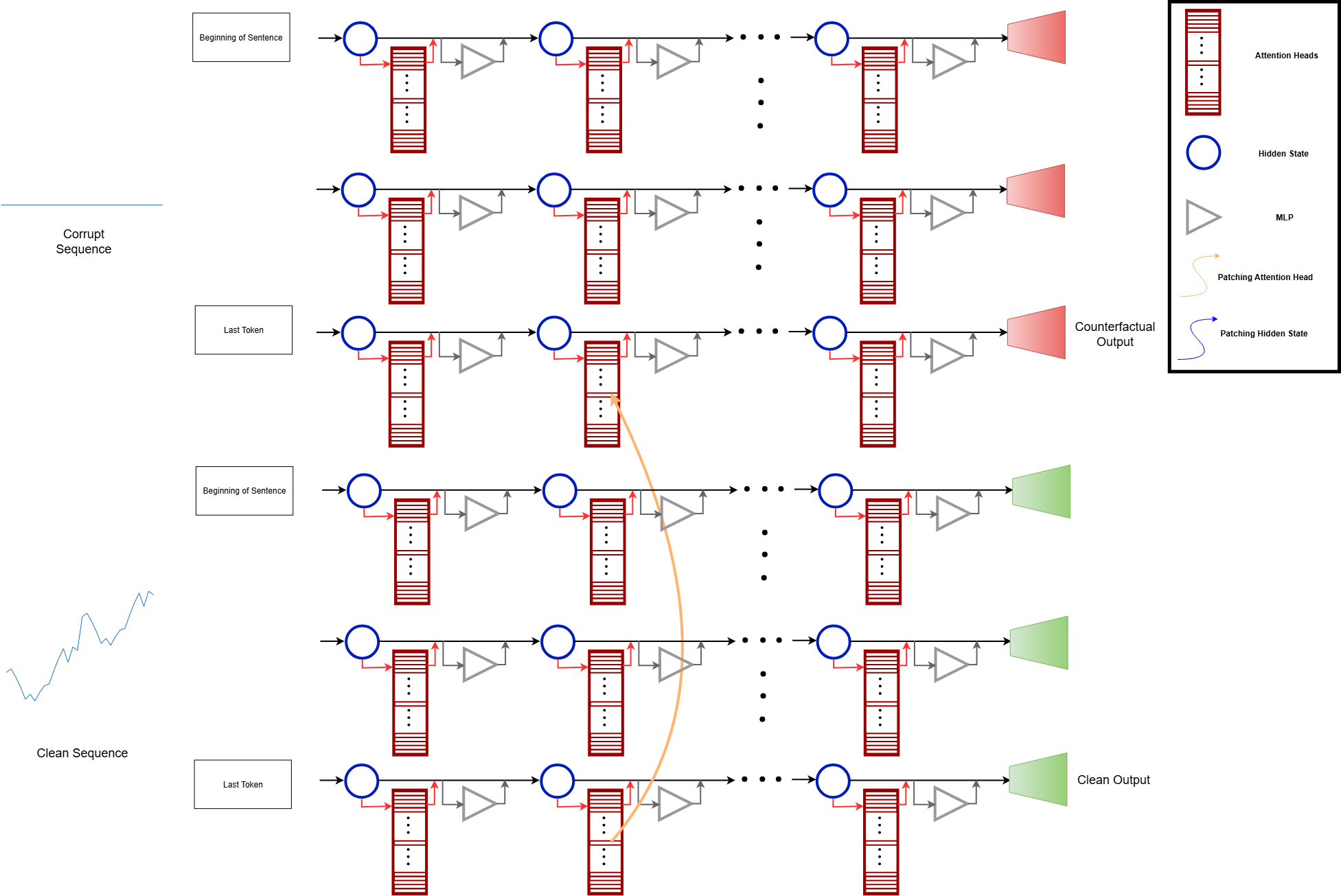} 
\caption{
Illustrates the head-level patching intervention that we use to isolate the functional role of individual attention heads in pattern recognition and delta retrieval. The clean sequence (bottom) contains structured first-difference patterns, while the corrupt sequence (top) is flat-valued. At each layer $l$, we patch only the output of a single attention head (at the final token position) from the clean forward pass into the corrupt one (shown with orange arrow), while all other heads and components remain unchanged. This targeted intervention tests whether the selected head causally contribute to the recognition of repeating structure, retrieval of a critical first difference, and contributes to accurate extrapolation in the final prediction. 
}

\label{fig:patchingheads}
\end{figure*}
\begin{figure*}[t]
\centering
\includegraphics[width=0.9\textwidth]{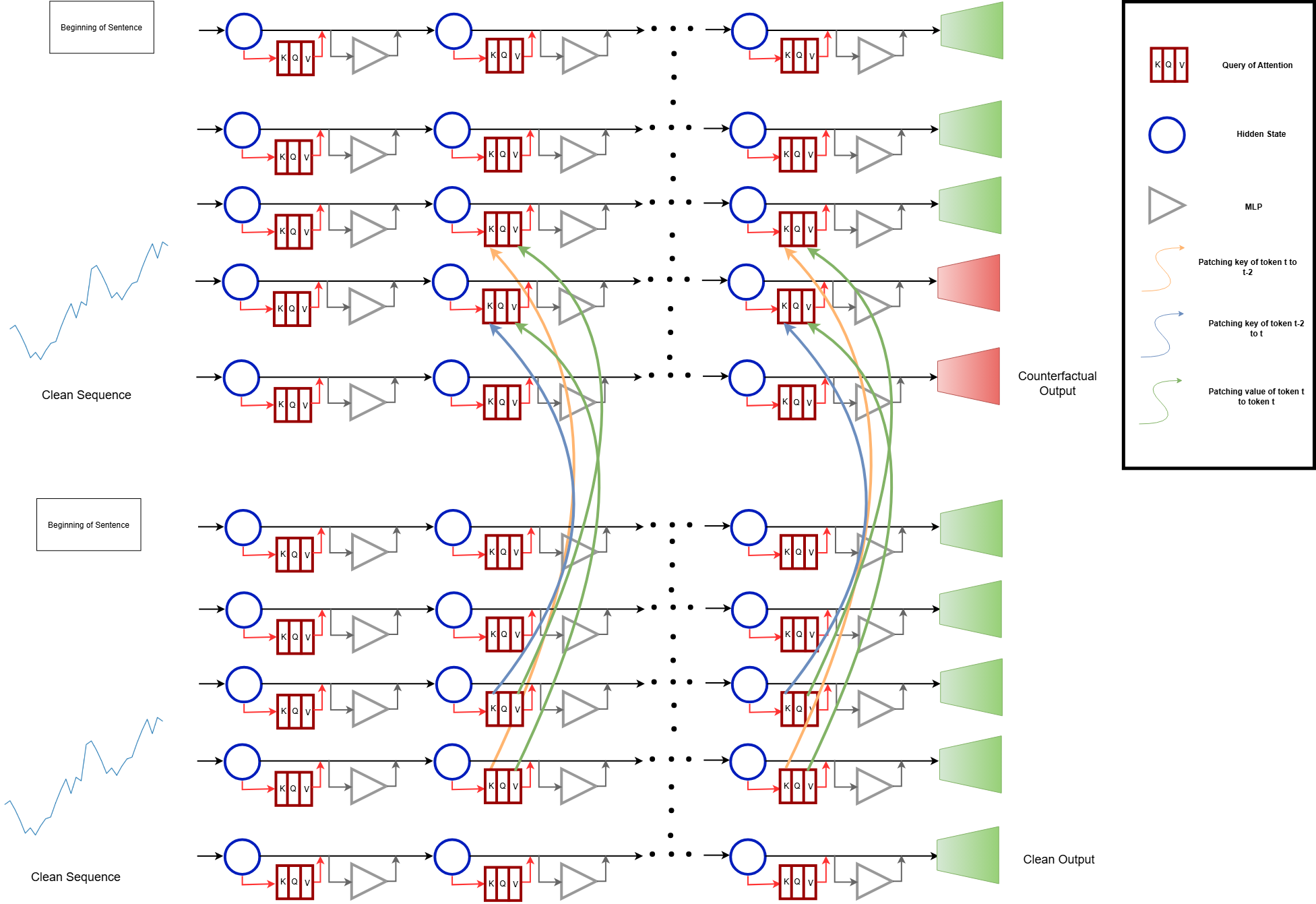} 
\caption{
Illustrates the key-swapping intervention that we use to disentangle the model’s reliance on phase (positional) cues versus delta-based representations during extrapolation. We swap keys from token positions 25 and 27—both identified as causally important—are swapped from layer $l$ onward (indicated by orange and gray arrows), while preserving the respective value vectors from layer $l$ onward. This setup tests whether the model selects the correct first difference for addition based on the original delta identity stored in the value vector or shifts its prediction in response to the redirected positional cue, thereby revealing its bias toward phase.
}

\label{fig:swapkeys}
\end{figure*}









\end{document}